\documentclass[10pt,twocolumn,letterpaper]{article}

\usepackage{iccv}              

\definecolor{iccvblue}{rgb}{0.21,0.49,0.74}
\usepackage[pagebackref,breaklinks,colorlinks,allcolors=iccvblue]{hyperref}

\usepackage{adjustbox}
\usepackage{multirow}
\usepackage{colortbl}
\usepackage[ruled,boxed]{algorithm2e}

\def\paperID{7075} 
\def\confName{ICCV}
\def\confYear{2025}

\title{You Only Click Once: Single Point Weakly Supervised 3D Instance Segmentation for Autonomous Driving}

\author{
    Guangfeng Jiang\textsuperscript{\rm 1},
    Jun Liu\textsuperscript{\rm 1}\thanks{Corresponding authors.},
    YongXuan Lv\textsuperscript{\rm 1},
    Yuzhi Wu\textsuperscript{\rm 1}, \\
    Xianfei Li\textsuperscript{\rm 2},
    Wenlong Liao\textsuperscript{\rm 2},
    Tao He\textsuperscript{\rm 2},
    Pai Peng\textsuperscript{\rm 2}\footnotemark[1]\\
    \textsuperscript{1}University of Science and Technology of China, \textsuperscript{2}COWAROBOT
}

\begin{document}
\maketitle
\begin{abstract}
Outdoor LiDAR point cloud 3D instance segmentation is a crucial task in autonomous driving. However, it requires laborious human efforts to annotate the point cloud for training a segmentation model. To address this challenge, we propose a \textbf{YoCo} framework, which generates 3D pseudo labels using minimal coarse click annotations in the bird's eye view plane. It is a significant challenge to produce high-quality pseudo labels from sparse annotations. Our YoCo framework first leverages vision foundation models combined with geometric constraints from point clouds to enhance pseudo label generation. Second, a temporal and spatial-based label updating module is designed to generate reliable updated labels. It leverages predictions from adjacent frames and utilizes the inherent density variation of point clouds (dense near, sparse far). Finally, to further improve label quality, an IoU-guided enhancement module is proposed, replacing pseudo labels with high-confidence and high-IoU predictions. Experiments on the Waymo dataset demonstrate YoCo’s effectiveness and generality, achieving state-of-the-art performance among weakly supervised methods and surpassing fully supervised Cylinder3D. Additionally, the YoCo is suitable for various networks, achieving performance comparable to fully supervised methods with minimal fine-tuning using only 0.8\% of the fully labeled data, significantly reducing annotation costs.
\end{abstract}
\section{Introduction}
\label{sec:intro}
3D point cloud segmentation (\eg, semantic segmentation, instance segmentation) is a fundamental research task in computer vision, particularly in the field of autonomous driving. In recent years, several studies~\cite{zhu2021cylindrical, hou2022point, xu2021rpvnet, kong2023rethinking, lai2023spherical, wu2024ptv3, wang2024sfpnet} have achieved promising results, 
largely attributable to advancements in neural network architectures~\cite{yan2018second, zhu2021cylindrical, zhao2021ptv1, wu2022ptv2, wu2024ptv3} and the emergence of high-quality autonomous driving datasets~\cite{caesar2020nuscenes, behley2019semantickitti, geiger2013kitti, sun2020waymo}. However, point cloud segmentation tasks typically rely on dense point-wise annotations, which are labor-intensive and costly. For example, in the ScanNet dataset~\cite{dai2017scannet}, it takes an average of 22.3 minutes to annotate a single scene. Consequently, reducing the reliance on dense point annotations is an economically beneficial yet challenging problem.

\begin{figure}[t]
    \centering
    \includegraphics[width=0.90\linewidth]{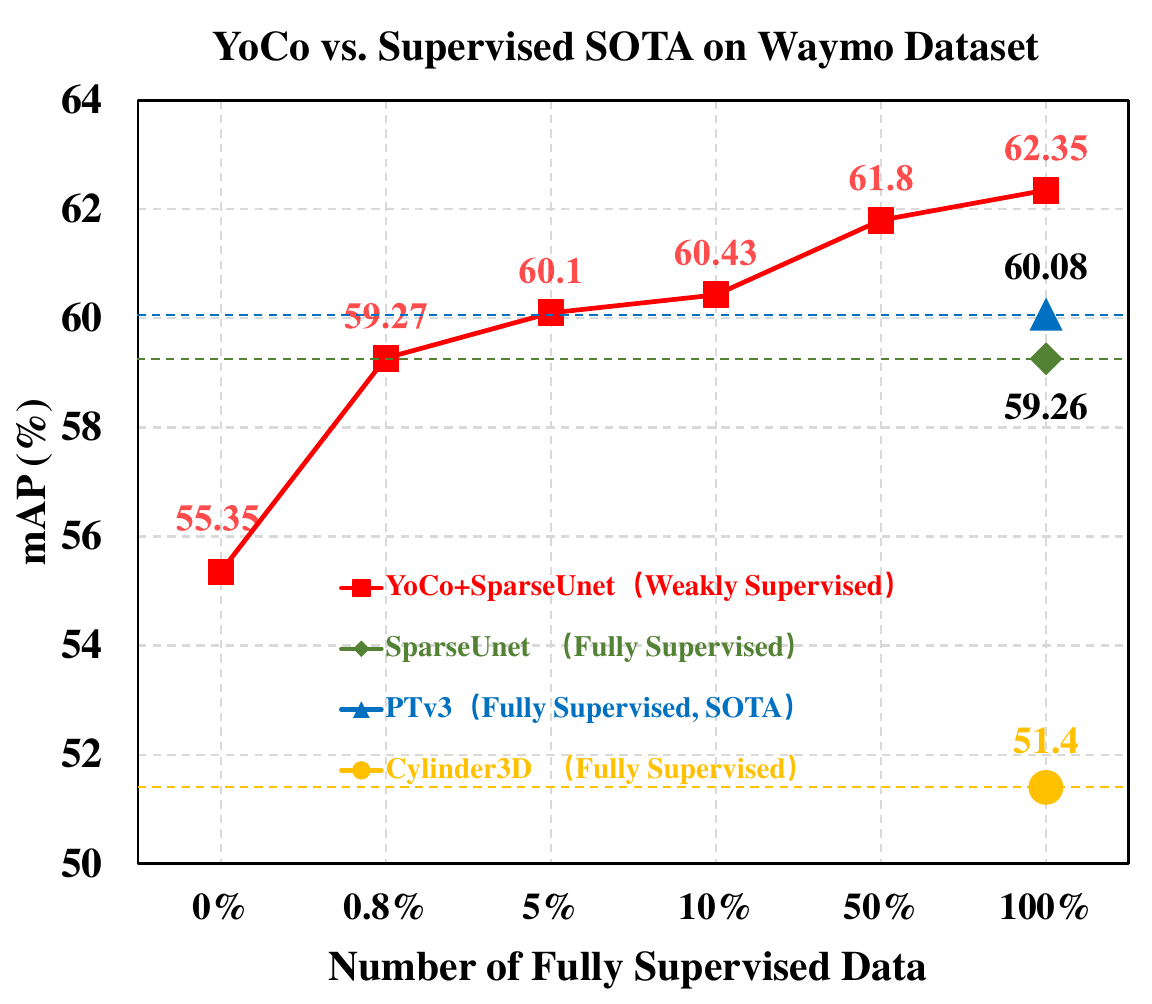}
    \vspace{-2mm}
    \caption{\textbf{3D Instance Segmentation Performance Comparison.} The weakly supervised YoCo for fine-tuning compared to fully supervised methods. The results show that our YoCo outperforms fully supervised Cylinder3D without fine-tuning (0\%). Fine-tuning YoCo with 0.8\% and 5\% labeled data exceeds the fully supervised SparseUnet and state-of-the-art (SOTA) PTv3, respectively.}
    \vspace{-4mm}
    \label{fig:finetune}
\end{figure}

Recent studies have attempted to address the problem of weakly supervised segmentation on 3D point clouds. Existing methods utilize various types of weak labels, such as sparse point-level labels~\cite{hu2022sqn, zhang2022not, chen2024foundation}, scribble-level labels~\cite{unal2022scribble, wang2024label}, and box-level labels~\cite{li2024box2mask, ngo2023gapro, yu20243when3dbox, jiang2024mwsis}. However, most of these approaches focus on semantic segmentation, while instance segmentation is more complex, as it requires distinguishing different instances within the same semantic category. For 3D instance segmentation tasks, the annotation of 3D bounding boxes is still expensive, although methods~\cite{li2024box2mask, ngo2023gapro, yu20243when3dbox} using 3D bounding boxes as weak supervision have achieved promising results. A recent work, MWSIS~\cite{jiang2024mwsis}, explores weakly supervised instance segmentation for outdoor LiDAR point clouds using low-cost 2D bounding boxes as weak supervision, but with a large performance gap with the fully supervised approach.
Inspired by the work mentioned above, we rethink whether there is a method with lower labeling cost to get better instance segmentation performance, and even narrow the gap between weakly supervised and fully supervised methods.

With this motivation, we propose a single-point supervised instance segmentation framework called \textbf{YoCo}. In this framework, a single click annotation per object in the bird's eye view (BEV) plane is sufficient to generate the corresponding 3D pseudo label for that object. As known, it is a non-trivial task to generate dense 3D pseudo labels from sparse click annotations. Following ~\cite{jiang2024mwsis}, click annotations can be used as prompts for SAM~\cite{kirillov2023segment} to generate corresponding 2D masks, which are then projected to obtain dense 3D pseudo labels. However, a major challenge lies in the limited zero-shot capability of SAM, resulting in noisy or inaccurate 2D masks. Therefore, a critical challenge lies in filtering high-quality 3D pseudo labels from these noisy outputs.
To address this challenge, we introduce a 3D pseudo label generation module based on the vision foundation models (VFMs), named VFM-PLG. Specifically, we use click annotations to obtain the corresponding 2D masks through VFMs, then leverage the geometric constraints (\eg, size, volume, depth, \etc) of the corresponding 3D masks to filter out high-quality 3D pseudo labels. 
In addition, to further improve the quality of pseudo labels, we take advantage of the generalization and robustness of neural networks by introducing two key modules: temporal and spatial-based label updating (TSU) module, and intersection-over-union (IoU)-guided label enhancement (ILE) module. The TSU module refines and updates the pseudo labels by incorporating predictions from adjacent frames, while the ILE module further enhances label quality by replacing lower-quality labels offline with more accurate predictions.

Experimental results show that our YoCo significantly outperforms previous state-of-the-art weakly supervised 3D instance segmentation methods and even surpasses fully supervised Cylinder3D~\cite{zhu2021cylindrical}, as shown in Figure \ref{fig:finetune}.
Additionally, YoCo demonstrates strong generality, making it suitable for various networks. Moreover, by fine-tuning the model with only 0.8\% of fully supervised data, it can surpass the performance of the fully supervised baseline.
In summary, our contributions are summarized as follows:
\begin{itemize}
    \item To the best of our knowledge, we are the first to propose using click annotations for instance segmentation of outdoor LiDAR point clouds. This method significantly reduces the burden of instance segmentation annotations.
    \item We propose the VFM-PLG, which combines VFMs with geometric constraints information of the object to generate high-quality pseudo labels. In addition, the TSU and ILE modules further improve pseudo label quality by leveraging the generalization and robustness of neural networks.
    \item Extensive experiments on the Waymo dataset demonstrate that our YoCo achieves state-of-the-art performance in weakly supervised instance segmentation, surpassing fully supervised methods such as Cylinder3D. Furthermore, YoCo exhibits strong generality, making it applicable across various networks.
\end{itemize}
\begin{figure*}[ht]
\centering
   \includegraphics[width=0.95\linewidth]{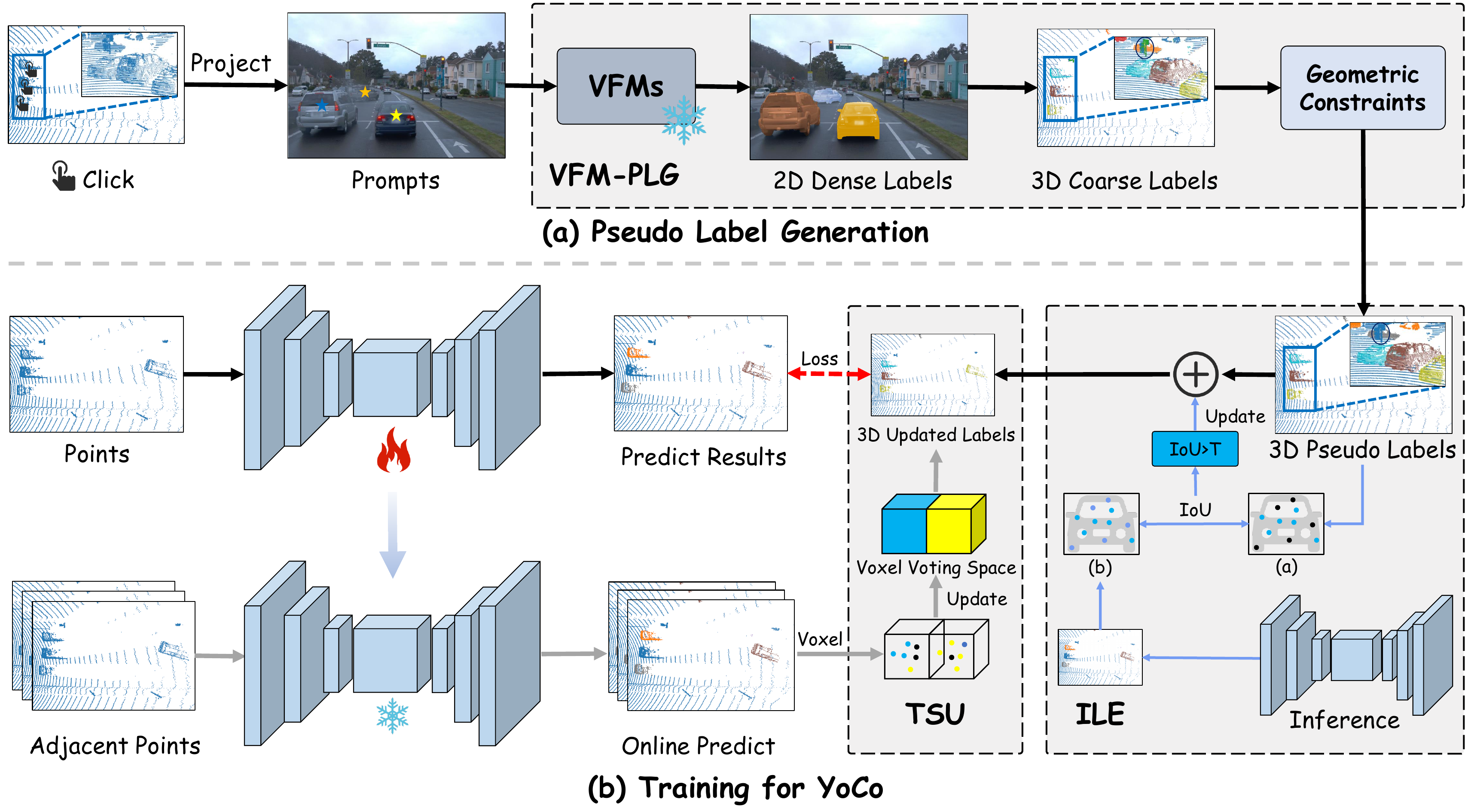}
   \vspace{-2mm}
   \caption{\textbf{Overview of YoCo Framework.} The YoCo consists of two main components: (a) pseudo label generation and (b) network training. For pseudo label generation, the VFM-PLG module produces high-quality pseudo labels using both VFMs and geometric constraints. For network training, our YoCo adopts the classic Mean Teacher~\cite{tarvainen2017meanteacher} structure. The TSU module performs online updates to the pseudo labels by using predictions from adjacent frames in a voxelized manner. Additionally, the ILE module enhances offline pseudo labels by leveraging high-confidence and high-IoU predictions to improve their quality.}
   \vspace{-2mm}
\label{fig:framework}
\end{figure*}
\section{Related Work}
\label{sec:related_work}
\noindent\textbf{LiDAR-Based Fully Supervised 3D Segmentation.}
Existing 3D LiDAR point cloud segmentation methods can be divided into three types according to data representation: point-based, projection-based, and voxel-based.

Point-based methods~\cite{qi2017pointnet, qi2017pointnet++, wu2019pointconv, thomas2019kpconv, zhao2021ptv1, wu2022ptv2, wu2024ptv3} directly use raw point clouds as the input. The classic PointNet~\cite{qi2017pointnet} utilizes the permutation invariance of both pointwise MLPs and pooling layers to aggregate features across a set. KPConv~\cite{thomas2019kpconv} and PointConv~\cite{wu2019pointconv} construct continuous convolution to directly process 3D points. The Point Transformer series~\cite{zhao2021ptv1, wu2022ptv2, wu2024ptv3} adopt the transformer architecture to extract features from 3D points.
Projection-based methods~\cite{ando2023rangevit, kong2023rangeformer, milioto2019rangenet++} project 3D points onto 2D images to form regular representations, allowing the use of well-established neural networks from 2D image processing. RangeViT~\cite{ando2023rangevit} directly applies a pre-trained ViT model as the encoder and fine-tunes it, demonstrating the feasibility of transferring 2D knowledge to 3D tasks. Rangeformer~\cite{kong2023rangeformer} and RangeNet++~\cite{milioto2019rangenet++} use an encoder-decoder hourglass-shaped architecture as the backbone for feature extraction.
Other methods~\cite {cheng20212-s3net, choy2019Minkowski, zhu2021cylindrical, graham20183d} convert point clouds into regular 3D voxelization. SSCN~\cite{graham20183d} introduces sparse convolutional networks to handle voxelized sparse point clouds. Cylinder3D~\cite{zhu2021cylindrical} introduces 3D cylindrical partitioning and asymmetrical 3D convolutions to handle the sparsity and varying density of outdoor point clouds.

While point-based methods deliver high performance, they come with significant computational costs due to the large-scale raw LiDAR data. On the other hand, projection-based methods are more efficient but lose valuable internal geometric information, resulting in suboptimal performance. Taking both time and memory efficiency into account, we adopt voxelized representations and select a sparse convolutional U-Net~\cite{shi2020PartA2} as our backbone.

\noindent\textbf{Weakly Supervised 3D Instance Segmentation.}
Point cloud segmentation has made significant progress in fully supervised settings. However, dense point-wise annotation is costly. To reduce the annotation burden, some work~\cite{wei2020mprm, hu2022sqn, zhang2022not, chen2024foundation, unal2022scribble, wang2024label,li2024box2mask, ngo2023gapro, yu20243when3dbox, jiang2024mwsis, guo2024sam} has explored using various weak supervision signals.


For 3D instance segmentation tasks, 3D bounding boxes provide coarse information about instance objects, making instance segmentation feasible. Box2Mask~\cite{li2024box2mask} is the first work to use 3D bounding boxes as weak supervision labels. GaPro~\cite{ngo2023gapro} proposes a gaussian process method to address pseudo label ambiguity in overlapping regions of multiple 3D bounding boxes. CIP-WPIS~\cite{yu20243when3dbox} leverages 2D instance knowledge and 3D geometric constraints to handle the 3D bounding box perturbation issues. Additionally, MWSIS~\cite{jiang2024mwsis} is the first work to use 2D bounding boxes as weak supervision signals for outdoor point cloud segmentation. It introduces various fine-grained pseudo label generation and refinement methods, and explores the possibility of integration with SAM~\cite{kirillov2023segment}. However, both 2D and 3D bounding boxes still involve considerable annotation costs. Our method only requires a click on the object in the BEV plane to generate high-quality pseudo labels.

\noindent\textbf{Click-Level Annotation for 3D Perception Tasks.}
Click-level annotation is a highly efficient and labor-saving labeling method. The recent work ~\cite{liu2021onethingoneclick, tao2022seggroup, liu2023clickseg, meng202ws3d, meng2021ws3d, zhang2023vitwss3d, xia2024oc3d} has begun to incorporate it into various 3D perception tasks.

One Thing One Click~\cite{liu2021onethingoneclick} employs click-level labels and introduces a graph propagation module to iteratively generate semantic pseudo labels. SegGroup~\cite{tao2022seggroup} propagates click-level labels to unlabeled segments through iterative grouping, generating instance pseudo labels. Meanwhile, ClickSeg~\cite{liu2023clickseg} presents a method that uses k-means clustering with fixed initial seeds to generate instance pseudo labels online. In the field of 3D object detection, WS3Ds~\cite{meng202ws3d, meng2021ws3d} annotate object centers in the BEV plane. It utilizes these center clicks as supervision signals to generate cylindrical proposals, and then employs a small amount of ground truth to train the network to produce 3D bounding boxes. ViT-WSS3D~\cite{zhang2023vitwss3d} proposes using a vision transformer to construct a point-to-box converter. 

The aforementioned studies, especially those focusing on weakly supervised 3D instance segmentation, primarily address indoor point clouds. In contrast, 3D instance segmentation of outdoor LiDAR point clouds remains largely unexplored. Although MWSIS~\cite{jiang2024mwsis} utilizes 2D bounding boxes, which incur lower annotation costs, it still exhibits a significant performance gap compared to fully supervised methods. To further minimize annotation costs, we propose a weakly supervised instance segmentation framework that relies solely on click-level annotations, effectively narrowing the gap with fully supervised approaches.

\section{Method}
\label{sec:method}
Our goal is to generate high-quality 3D instance pseudo labels using sparse click-level annotations, and to narrow the performance gap between weakly supervised and fully supervised methods. To achieve this, we propose a simple yet effective framework, \textbf{YoCo}, which integrates pseudo label generation with network training, as illustrated in Figure \ref{fig:framework}. By leveraging the minimal input of click annotations, YoCo efficiently creates reliable pseudo labels that maintain strong performance, even with limited supervision. The detailed process is outlined as follows:

For pseudo label generation in Figure \ref{fig:framework}(a), given a set of calibrated images and point cloud data, we first annotate the point cloud with click-level labels in the BEV plane. These labels are then projected onto the corresponding images, and processed by our proposed VFM-PLG module. The VFM-PLG leverages the VFMs and geometric constraints from the point cloud to generate high-quality 3D pseudo labels, as described in Section \ref{sec:module_1}.

For network training in Figure \ref{fig:framework}(b), we adopt the MeanTeacher~\cite{tarvainen2017meanteacher} method, which involves a student network and a teacher network. The teacher network is updated using an exponential moving average (EMA) of the student’s weights. It predicts labels from adjacent frames, and the TSU module uses these predictions to refine the 3D pseudo labels generated by the VFM-PLG. This refinement incorporates temporal and spatial information from adjacent frames, as detailed in Section \ref{sec:module_2}. 

Additionally, to further boost the reliability of the pseudo labels, we introduce the ILE module. This module enhances the labels offline by using high-confidence and high-IoU results to update the 3D pseudo labels, further improving the performance of our method, as outlined in Section \ref{sec:module_3}.

\subsection{Preliminary}
\label{sec:prelimiinary}
Given a set of calibrated images and point cloud data, we utilize sensor calibration to project the point cloud onto the images, establishing a mapping relationship between the 3D points and image pixels. Specifically, consider a set of points $\mathbf{P}^{3\mathbf{d}}=\left\{ p_{i}^{3d} \right\} _{i=1}^{N}=\left\{ \left( x_i,y_i,z_i \right) \right\} _{i=1}^{N} \in \mathbb{R}^{N\times 3}$ we can obtain the corresponding pixel coordinates $\mathbf{P}^{2\mathbf{d}}=\left\{ p_{i}^{2d} \right\} _{i=1}^{N}=\left\{ \left( u_i,v_i \right) \right\} _{i=1}^{N} \in \mathbb{R}^{N\times 2}$ by applying the projection transformation formula:
\begin{equation}
\label{eq:project}
    z_c\left( u_i,v_i,1 \right) ^T=\mathbf{M}_{int}\times \mathbf{M}_{ext}\times \left( x_i,y_i,z_i,1 \right) ^T
\end{equation}
where $N$ is the number of the points, $z_c$ represents the depth of the point in the camera coordinate system, $\mathbf{M}_{int}$ and $\mathbf{M}_{ext}$ denote the intrinsic and extrinsic parameters of the camera, respectively.

\begin{figure}[ht]
    \centering
    \includegraphics[width=0.9\linewidth]{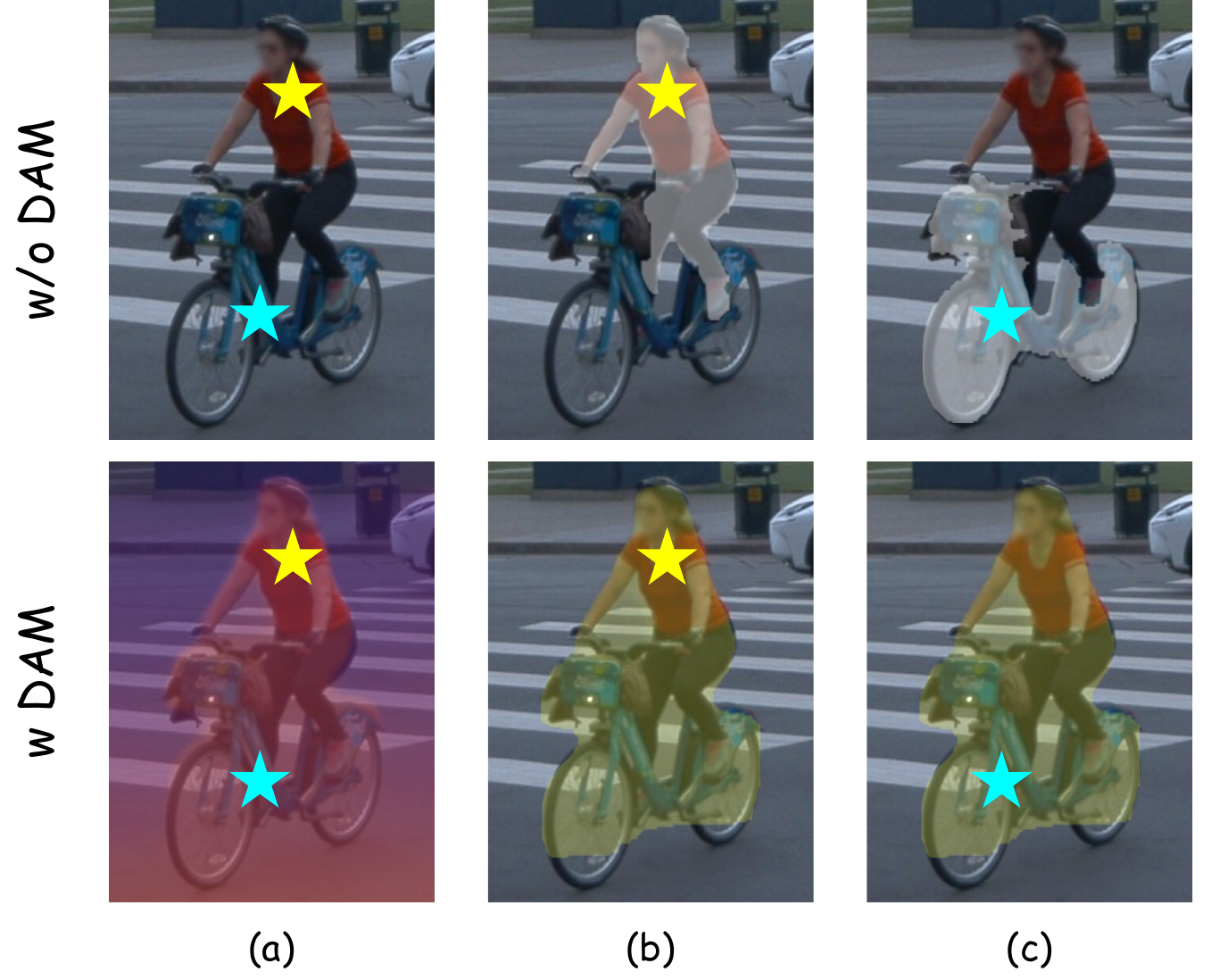}
    \vspace{-2mm}
    \caption{\textbf{Our VFM-PLG for the Composite Categories.} Comparison of 2D mask results for the same instance with different prompts (colored stars). \textit{w/o} shows results without DAM, while \textit{w} shows results with DAM. Using the DAM model yields more consistent and accurate results across different prompts.}
    \vspace{-1mm}
    \label{fig:sam4cyc}
\end{figure}

\subsection{VFM-Based Pseudo Label Generation}
\label{sec:module_1}
SAM~\cite{kirillov2023segment} is a vision foundation model that inputs images and prompts to generate corresponding 2D masks. The prompts include points, bounding boxes, masks, and texts. By leveraging the projection relationship described in Equation \ref{eq:project}, we project the click-level annotations from point clouds onto the image as prompts to obtain the object's 2D masks. Points that fall within these 2D mask regions can be considered as 3D pseudo labels. This process can be formally expressed by the following equation:
\begin{equation}
\label{eq:color}
    m_i = Color(SAM(I_i,p_i,c_i))
\end{equation}
where $m_i$ denotes the 3D pseudo label to the $i$-th click annotation, \textit{Color} represents the operation of mapping a 2D mask to a 3D pseudo label, $I_i$, $p_i$, and $c_i$ represent the image, 2D coordinate, and the class label corresponding to the $i$-th click annotation, respectively.

However, this approach faces three challenges. First, SAM struggles with segmenting composite categories like cyclists. The second one is inaccuracies in the SAM segmentation mask and the projection relationship. Finally, due to the lack of height information in the BEV plane, there may be multiple corresponding 3D points in the current click, and an incorrect prompt will result in an inaccurate segmentation result.

\begin{figure}[ht]
    \centering
    \includegraphics[width=1.0\linewidth]{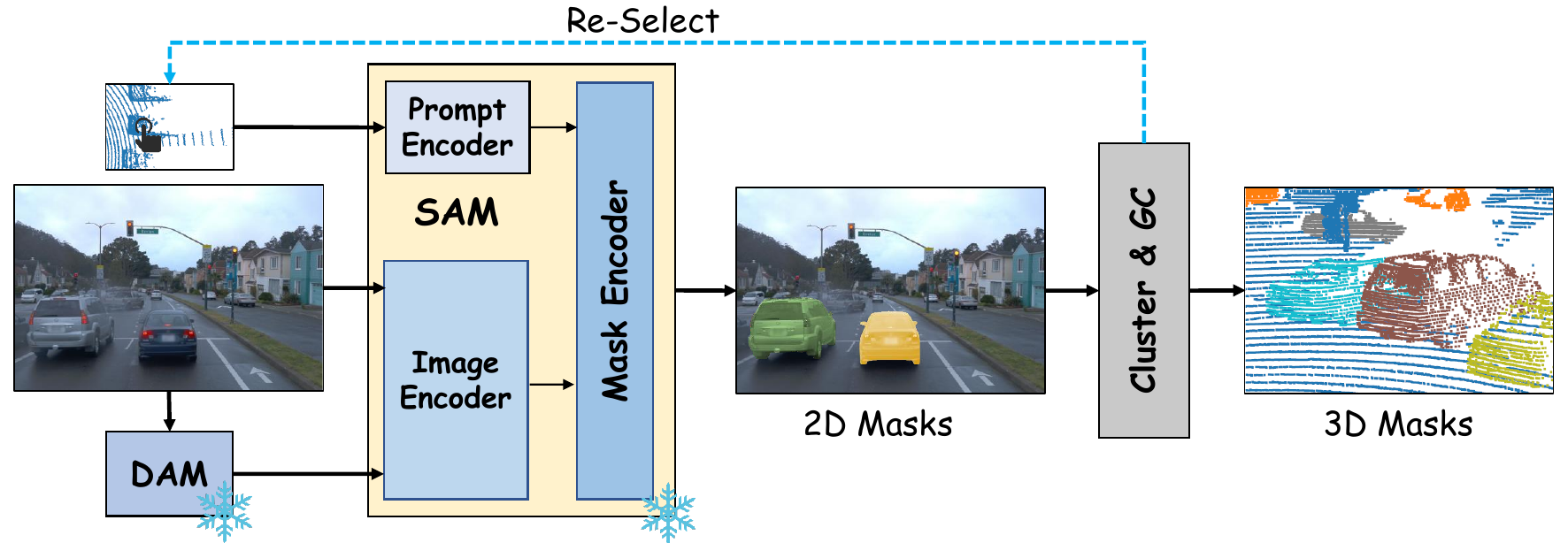}
    \vspace{-2mm}
    \caption{\textbf{Overview of VFM-PLG Module.} The blue dashed line indicates that if the generated 3D mask does not satisfy geometric constraints, another point is selected as the prompt. \textit{GC} denotes that the point cloud is processed using geometric constraints.}
    \label{fig:vfm-plg}
    \vspace{-3mm}
\end{figure}

To address the first issue, we utilize the Depth Anything Model (DAM)~\cite{yang2024depthanythin} as an auxiliary tool to perform depth-based smoothing, particularly for composite categories like cyclists (as shown in Figure \ref{fig:sam4cyc}).
Specifically, the image is processed through the DAM to generate a depth map, and the depth map's features are then used to interact with the prompt's features, resulting in the corresponding 2D mask. (as shown in Figure \ref{fig:vfm-plg}). Therefore, Equation \ref{eq:color} is updated as:
\begin{equation}
\label{eq:color2}
    m_i= Color\left( SAM\left( DAM\left( I_i \right) ,p_i,c_i \right) \right)
\end{equation}

As for the last two issues, point cloud geometric constraints are proposed for filtering the labels. Specifically, we first obtain a 3D pseudo label $m_i$ for the $i$-th object through the projection relationship. Then a clustering algorithm is applied to $m_i$, resulting in a set of clusters. The cluster containing the click annotation $p_i$ is identified as the 3D pseudo label for that object ($Find$ operation in Equation \ref{eq:filter}).
Next, a geometric consistency check is performed on the identified cluster ($Filter$ operation in Equation \ref{eq:filter}). The label is retained if the cluster satisfies certain geometric conditions; otherwise, the pseudo label is discarded, \ie, $m_i=\left\{0\right\}^N_{i=1}$. This process can be represented as follows:

\begin{equation}
\label{eq:filter}
    \tilde{m}_i=Filter\left( Find\left( Cluster\left( m_i \right) \right) \right)
\end{equation}

If the current click corresponds to multiple points, we iterate to select one as the prompt. If the current result meets the geometric constraints, it is retained as the 3D pseudo label; otherwise, a new point is re-selected as the prompt to generate the pseudo label, as indicated by the blue dashed line in Figure \ref{fig:vfm-plg}.


\subsection{Temporal and Spatial-Based Label Updating}
\label{sec:module_2}
To improve the quality of pseudo labels generated by the VFM-PLG module, we propose the temporal and spatial-based label updating module. This module leverages the generalization of neural networks by using high-reliability predictions from adjacent frames to update the pseudo labels of the current frame online.

To transform the point clouds from adjacent frames to the current frame, a coordinate system transformation is required, which can be expressed by the following equation:
\begin{equation}
\label{eq:transform}
    \mathbf{P}_t=\mathbf{T}_{t}^{-1}\times \mathbf{T}_{adj}\times \mathbf{P}_{adj}
\end{equation}
where $\mathbf{T}_{t}$ is the ego-car pose in the current frame, $\mathbf{T}_{adj}$ is the ego-car pose in the adjacent frame, $\mathbf{P}_t$ and $\mathbf{P}_{adj}$ correspond to the coordinates of the point cloud in the current frame and the adjacent frame, respectively.

Unlike MWSIS's~\cite{jiang2024mwsis} PVC module, our method does not require establishing a voting space from the previous training epochs, which reduces memory requirements during training. Moreover, since it is difficult to establish a one-to-one correspondence between point clouds of adjacent frames, we employ a voxel voting mechanism. An online voxel voting space $S$ is set up, where each voxel updates its corresponding label according to a predefined update strategy. The current frame requires voxelization of the point cloud, and the updated labels are obtained from the corresponding voxel space. The specific update strategy is as follows:
\begin{itemize}
    \item \textbf{Soft voting strategy.} Consider a voxel that contains $n$ points, where each point $p_i^v$ has an associated classification confidence scores $s^v_i \in \mathbb{R}^{num}$, where the $num$ denotes the class number. We average the classification confidence scores for all points within the voxel, and then identify the class $c^v_i$ with the highest score. If this score exceeds a set threshold $T_{s}$, the class is assigned as the voxel's label. The above process can be represented by Equation \ref{eq:softvoting}. Unlike the method of directly selecting the class with the most points, this approach enhances robustness against prediction noise.
    \begin{equation}
    \label{eq:softvoting}
        c_{i}^{v}=\begin{cases}
        	\text{arg}\max \bar{s}&		\text{if\,\,}\max \bar{s} >T_s\\
        	-1&		\text{otherwise}\\
        \end{cases}
    \end{equation}
    where $\bar{s}= \frac{1}{n}\sum_{i=1}^{n}{s_{i}^{v}}$, $-1$ denotes the ignored label.
    \item \textbf{Distance-based reliability update strategy.} To improve the reliability of voting labels, we consider that a greater number of voxel points and higher point confidence result in more reliable voting. Given that LiDAR point clouds are dense near the sensor and sparse farther away, we dynamically adjust the voting threshold: voxels closer to the sensor require more votes and higher confidence for label assignment.
\end{itemize}
By applying the aforementioned update strategy, a reliable voxel voting space $S$ is constructed, which is then used to update the labels of the current frame. For further details, refer to Algorithm \ref{alg:tsu} in the supplementary material.

\begin{table*}[ht]
  \centering
  \adjustbox{max width=1.0\textwidth}{
    \begin{tabular}{ccccccccccc} 
    \toprule
    \multirow{2}{*}{Supervision} & \multirow{2}{*}{Annotation} & \multirow{2}{*}{Model} & \multicolumn{4}{c}{3D Instance Segmentation (AP)}        & \multicolumn{4}{c}{3D Semantic Segmentation (IoU)}            \\ 
    \cline{4-11}
                                 &                             &                        & mAP   & Veh.  & Ped.  & Cyc.  & mIoU   & Veh.   & Ped.   & Cyc.    \\ 
    \hline
    \multirow{3}{*}{Full}        & 3D Mask                     & Cylinder3D~\cite{zhu2021cylindrical}             & 51.40 & 75.31 & 38.12 & 40.76 & 78.903 & 96.476 & 83.666 & 56.567  \\
                                 & 3D Mask                     & PTv3~\cite{wu2024ptv3}                   & 60.08 & 75.73 & 53.63 & 51.32 & 83.679 & 96.686 & 85.500 & 68.852  \\
                                 & 3D Mask                     & SparseUnet~\cite{shi2020PartA2}             & 59.26 & 80.25 & 56.95 & 40.59 & 79.505 & 96.675 & 81.906 & 59.933  \\ 
    \arrayrulecolor{black}\cline{1-1}\arrayrulecolor{black}\cline{2-11}
    \multirow{9}{*}{Weak}        & 3D Box                      & SparseUnet~\cite{shi2020PartA2}             & 49.32 & 69.00 & 45.96 & 33.01 & 72.545 & 89.471 & 73.581 & 54.582  \\
                                 & 2D Box                      & SparseUnet~\cite{shi2020PartA2}             & 35.48 & 44.54 & 36.84 & 25.08 & 63.831 & 74.102 & 72.113 & 45.278  \\
                                 & 2D Box$^*$                      & SparseUnet~\cite{shi2020PartA2}             & 45.12 & 64.06 & 40.06 & 31.23 & 75.571 & 93.418 & 77.982 & 55.312  \\
                                 & 2D Box                      & MWSIS~\cite{jiang2024mwsis}                  & 48.42 & 61.45 & 45.23 & 38.59 & 75.898 & 90.369 & 78.996 & 58.329  \\
                                 & Click$^*$                       & SparseUnet~\cite{shi2020PartA2}             & 40.19 & 59.15 & 36.98 & 24.45 & 67.510 & 76.225 & 76.118 & 50.187  \\ 
    \cline{2-11}
                                 & Click$^\dag$                       & Cylinder3D~\cite{zhu2021cylindrical}             & 45.12 & 60.72 & 36.52 & 38.11 & 70.068 & 76.653 & \underline{79.123} & 54.428  \\
                                 & Click$^\dag$                       & PTv3~\cite{wu2024ptv3}                   & 46.59 & 62.59 & 38.12 & \underline{39.06} & \underline{72.776} & \textbf{82.109} & 77.056 & \underline{59.163}  \\
                                 & Click$^\dag$                      & SparseUnet~\cite{shi2020PartA2}             & \underline{47.37} & \underline{64.10} & \underline{41.50} & 36.51 & 72.189 & 79.850 & 78.619 & 58.097  \\ 
    \cline{2-11}
                                 & Click$^\dag$\cellcolor{lightgray}  & \textbf{YoCo (Ours)}\cellcolor{lightgray} & \textbf{55.35}\cellcolor{lightgray} & \textbf{67.69}\cellcolor{lightgray} & \textbf{55.25}\cellcolor{lightgray} & \textbf{43.12}\cellcolor{lightgray} & \textbf{74.770}\cellcolor{lightgray} & \underline{81.136}\cellcolor{lightgray} & \textbf{81.716}\cellcolor{lightgray} & \textbf{64.459}\cellcolor{lightgray}  \\
    \bottomrule
    \end{tabular}
  }
  \caption{\textbf{Performance Comparisons of 3D Instance and Semantic Segmentation on Waymo Validation Dataset.} \textbf{Bold} indicates optimal performance, and \underline{underline} indicates sub-optimal performance. $^*$ represents the pseudo label generated by SAM using the corresponding annotation as the prompt. $^\dag$ denotes the pseudo label generated by our VFM-PLG module. Abbreviations: vehicle (Veh.), pedestrian (Ped.), cyclist (Cyc.). }
  \vspace{-1mm}
  \label{tab:main results}
\end{table*}

\subsection{IoU-Guided Label Enhancement}
\label{sec:module_3}
To further leverage the robustness of neural networks and correct the erroneous pseudo labels generated by the VFM-PLG, we introduce an IoU-guided label enhancement module. This module performs an offline update of pseudo labels using high-confidence scores and high-IoU value predictions. Additionally, we adjust the confidence score threshold accordingly to accommodate the characteristic density variation in LiDAR point clouds (dense near, sparse far). The process can be represented as follows:
\begin{equation}
    m'_i=\begin{cases}
        \text{arg}\max \left( IoU_i \right)&		\,\,\text{if\,\,}s_i\ge T_{s_2}\,\,\text{and\,\,}IoU_i\ge T_{IoU}\\
        \tilde{m}_i&		\,\,\text{otherwise\,\,}\\
    \end{cases}
\end{equation}
where $IoU_i=\frac{\tilde{m}_i\cap \hat{m}_i}{\tilde{m}_i\cup \hat{m}_i}$, $m'_i$ and $\hat{m}_i$ represent the updated labels and the predicted labels, respectively, while $T_{s_2}$ and $T_{IoU}$ correspond to the predefined confidence scores threshold and IoU threshold.

\subsection{Loss}
\label{sec:loss}
The overall loss function of the YoCo is defined as:
\begin{equation}
    L = \alpha_1 L_{cls} + \alpha_2 L_{vote}
\end{equation}
where $L_{cls}$ denotes the cross entropy loss or focal loss, and $L_{vote}$ represents the L1 loss, $\alpha_1$ and $\alpha_2$ are hyperparameters to balance loss terms.

\section{Experiments}
\label{sec:experiment}
\subsection{Waymo Open Dataset}
Following the state-of-the-art MWSIS~\cite{jiang2024mwsis}, we conduct our experiments on version 1.4.0 of the Waymo Open Dataset (WOD)~\cite{sun2020waymo}, which includes both well-synchronized and aligned LiDAR points and images. The WOD consists of 1,150 sequences (over 200K frames), with 798 sequences for training, 202 sequences for validation, and 150 sequences for testing. For the 3D segmentation task, the dataset contains 23,691 and 5,976 frames for training and validation, respectively. We specifically focus on the vehicle, pedestrian, and cyclist categories for evaluation.

\subsection{Implementation Details}
\noindent\textbf{Click Setting.} For the click annotation, we use the average coordinates of each instance from the BEV plane as a reference and then select the nearest point to simulate the manual click. Meanwhile, in Table \ref{tab:click}, we also simulate the results with the manual annotation error.

\noindent\textbf{Evaluation Metric.} We adopt the same evaluation metrics as~\cite{jiang2024mwsis}. For 3D instance segmentation, we use average precision (AP) across different IoU thresholds to assess performance, while for 3D semantic segmentation, we use mean intersection-over-union (mIoU) as the evaluation metric.

\noindent\textbf{Training Setting.} 
We choose several classic backbones, including Cylinder3D~\cite{zhu2021cylindrical}, SparseUnet~\cite{shi2020PartA2}, and Point Transformer V3 (PTv3)~\cite{wu2024ptv3}, and use two separate heads: one for predicting semantic masks and another for grouping pixels into instances. Cylinder3D, PTv3, and SparseUnet are trained for 40, 50, and 24 epochs, respectively. All models are trained on 4 NVIDIA 3090 GPUs with a batch size of 8, using the AdamW~\cite{loshchilov2017adamw} optimizer.

\subsection{Results on WOD}

We compare the YoCo with other weakly supervised and fully supervised methods for 3D instance segmentation. Considering both computational time and memory efficiency, we select SparseUnet~\cite{shi2020PartA2} as the baseline for our experiments. Additionally, Table \ref{tab:yoco4other} presents the results of YoCo with different networks, further demonstrating the generality of our method. In Table \ref{tab:main results}, our YoCo achieves the best performance among weakly supervised methods. It outperforms the Click$^*$-based approach with a 15.16\% improvement in mAP and surpasses the state-of-the-art method MWSIS by 6.93\% mAP, while utilizing more cost-effective sparse click annotations. Additionally, compared with methods using 3D bounding boxes, which have higher annotation costs, our approach achieves a 6.03\% mAP improvement. Moreover, our method outperforms fully supervised Cylinder3D by 3.95\% mAP.

We also provide metrics for 3D semantic segmentation. Compared to the Click$^*$-based method, our approach achieves a 7.260\% improvement in mIoU. When compared to methods based on 3D bounding boxes, it achieves a 2.225\% mIoU increase. Additionally, our method reaches 94.76\% of the fully supervised performance.


\begin{table}[ht]
    \centering
    \adjustbox{max width=0.75\linewidth}{
    \begin{tabular}{ccccc} 
    \toprule
    \multicolumn{3}{c}{Method}                 & \multirow{2}{*}{mAP}      & \multirow{2}{*}{mIoU}       \\ 
    \cline{1-3}
    VFM-PLG          & TSU          & ILE          &                           &                             \\ 
    \hline
    $\mbox{-}$   & $\mbox{-}$   & $\mbox{-}$   & \multicolumn{1}{l}{25.36} & \multicolumn{1}{l}{51.919}  \\
    \hline
    $\checkmark$ & $\mbox{-}$   & $\mbox{-}$   & 47.37                     & 72.189                      \\
    $\checkmark$ & $\checkmark$ & $\mbox{-}$   & 52.18                     & 74.449                      \\
    $\checkmark$\cellcolor{lightgray} & $\checkmark$\cellcolor{lightgray} & $\checkmark$\cellcolor{lightgray} & \textbf{55.35}\cellcolor{lightgray}            & \textbf{74.770}\cellcolor{lightgray}             \\
    \bottomrule
    \end{tabular}
    }
    \caption{\textbf{Ablation Study of All Modules.}}
    \label{tab:ile}
    \vspace{-4mm}
\end{table}

\subsection{Ablation Study and Analysis}
\label{sec:ablation}

\noindent\textbf{Effect of all modules.}
Table \ref{tab:ile} presents the ablation study of all proposed modules. When solely the VFM-PLG module is utilized, we observe a substantial performance improvement, with instance segmentation and semantic segmentation improving by 22.01\% mAP and 20.270\% mIoU, respectively. This demonstrates the effectiveness of generating pseudo labels by combining the VFM module with geometric constraints from point clouds. When the TSU module is introduced, the performance further improves by 4.81\% mAP and 2.26\% mIoU. This highlights the value of leveraging the neural network’s generalization by using high-confidence predictions from adjacent frames to update the current frame’s labels,  improving the quality of the pseudo labels. Moreover, our proposed ILE module leverages the robustness of the network to perform offline refinement of the pseudo labels generated by the VFM-PLG. This approach leads to additional gains in label quality, with instance segmentation improving by 3.17\% mAP and semantic segmentation by 0.321\% mIoU. These results demonstrate the effectiveness of our framework in progressively refining pseudo labels and narrowing the performance gap between weakly supervised and fully supervised methods.

\noindent\textbf{Effect of the VFM-PLG.}
In Table \ref{tab:plg}, we conduct an ablation study to evaluate the impact of each component in the VFM-PLG module. The second row is the baseline performance, where the model is trained using labels generated from click annotations and the SAM. When a clustering algorithm is adopted to refine pseudo labels, the mAP improves substantially by 14.83\%. Additionally, applying the DAM to handle composite categories, such as cyclists, can further improve performance by 5.82\% AP. When geometric constraints are incorporated, the mAP reaches 47.37\%, surpassing methods that utilize 2D boxes as prompts. Combining these methods not only improves segmentation performance but also significantly reduces annotation costs, demonstrating the efficiency and practicality of our approach in weakly supervised 3D instance segmentation.
\begin{table}[ht]
    \centering
    \adjustbox{max width=0.7\linewidth}{
    \begin{tabular}{cccc}
    \toprule
    \multirow{2}{*}{Annotation~} & \multirow{2}{*}{Method} & \multicolumn{2}{c}{AP}           \\ 
    \cline{3-4}
                                 &                         & mAP            & Cyc.            \\ 
    \hline
    2D Box                       & SAM                     & 45.12          & 31.23           \\ 
    \hline
    \multirow{4}{*}{Click}       & SAM                     & 25.36          & 16.42           \\
                                 & + Cluster               & 40.19          & 24.45           \\
                                 & + DAM                   & 42.13          & 30.27           \\
                                 & + GC\cellcolor{lightgray}                    & \textbf{47.37}\cellcolor{lightgray} & \textbf{36.51}\cellcolor{lightgray}  \\
    \bottomrule
    \end{tabular}
    }
    \caption{\textbf{Ablation Study of VFM-PLG Module.}}
    \label{tab:plg}
    \vspace{-2mm}
\end{table}

\begin{table}[ht]
    \centering
    \adjustbox{max width=0.45\linewidth}{
    \begin{tabular}{ccc} 
    \toprule
    Frames & mAP            & mIoU             \\ 
    \hline
    0      & 47.37          & 72.189           \\ 
    \hline
    1      & 47.52          & 68.346           \\
    2\cellcolor{lightgray}      & \textbf{52.18}\cellcolor{lightgray} & \textbf{74.449}\cellcolor{lightgray}  \\
    3      & 51.07          & 74.107           \\
    5      & 48.17          & 72.345           \\
    \bottomrule
    \end{tabular}
    }
    \caption{\textbf{Ablation Study of the Number of Adjacent Frames.}}
    \label{tab:tsu1}
    \vspace{-1mm}
\end{table}

\begin{table}[ht]
    \centering
    \adjustbox{max width=0.9\linewidth}{
    \begin{tabular}{cccccc} 
    \toprule
    \multirow{2}{*}{Frames} & \multirow{2}{*}{Vote Mode} & \multicolumn{2}{c}{Dynamic} & \multirow{2}{*}{mAP} & \multirow{2}{*}{mIoU}  \\ 
    \cline{3-4}
                            &                            & Count        & Score        &                      &                        \\ 
    \hline
    0                       & $\mbox{-}$                 & $\mbox{-}$   & $\mbox{-}$   & 47.37                & 72.189                 \\ 
    \hline
    \multirow{5}{*}{2}      & \multirow{2}{*}{Hard}      & $\mbox{-}$   & $\mbox{-}$   & 48.54                & 73.138                 \\
                            &                            & $\checkmark$ & $\mbox{-}$   & 48.68                & 73.215                 \\ 
    \cline{2-6}
                            & \multirow{3}{*}{Soft}      & $\mbox{-}$   & $\mbox{-}$   & 49.97                & 73.152                 \\
                            &                            & $\checkmark$ & $\mbox{-}$   & 50.83                & 73.564                 \\
                            &                            & $\checkmark$\cellcolor{lightgray} & $\checkmark$\cellcolor{lightgray} & \textbf{52.18}\cellcolor{lightgray}       & \textbf{74.449}\cellcolor{lightgray}        \\
    \bottomrule
    \end{tabular}
    }
    \caption{\textbf{Ablation Study of Voting Strategy in the TSU Module.} \textit{Hard} voting means that the predicted label of the majority class among the points within a voxel is selected as the updated label. In contrast, \textit{Soft} voting assigns the updated label based on the class with the highest average confidence score across all points within the voxel. \textit{Count} and \textit{Score} represent the use of dynamic thresholds, \textit{Count} dynamically adjusts the required number of votes, while \textit{Score} dynamically adjusts the confidence score threshold for label updates.}
    \label{tab:tsu2}
    \vspace{-2mm}
\end{table}

\noindent\textbf{Effect of the TSU.}
Tables \ref{tab:tsu1} and \ref{tab:tsu2} provide the ablation studies for the TSU module. In Table \ref{tab:tsu1}, we analyze the impact of using different numbers of frames for label updates. We observe a steady improvement in performance as the number of frames increases up to a certain point. Specifically, when updating labels using predictions from two adjacent frames, the method achieves its best performance, with 52.18\% mAP and 74.449\% mIoU. However, as the number of frames exceeds three, the increased motion of dynamic objects causes greater discrepancies between frames, leading to erroneous votes and a consequent drop in overall performance.

Table \ref{tab:tsu2} compares several voting strategies to assess their impact on segmentation results. When employing the hard voting method (row 3), compared to the baseline method (row 1), we observe only modest improvements, with gains of 1.17\% mAP and 0.949\% mIoU. This limited improvement suggests that the hard voting is not sufficiently robust to handle noise in the predictions. In contrast, when applying the soft voting method (row 4), the performance improvement is more pronounced, yielding 2.6\% mAP and 0.963\% mIoU gains.

Furthermore, when we integrate our distance-based reliability update strategy (row 6), which adapts to the varying density of LiDAR point clouds based on their proximity to the sensor, we observe a significant increase in performance. Specifically, instance segmentation improves by 4.81\% mAP, while semantic segmentation sees a gain of 2.260\% mIoU. These results highlight the robustness of our proposed strategy in leveraging the inherent properties of LiDAR point clouds, particularly the denser distribution of points near the sensor and the sparser distribution at greater distances, to enhance the reliability of pseudo labels during training. This approach not only improves segmentation accuracy but also addresses challenges posed by noisy or uncertain predictions.


\noindent\textbf{Effect of the manual annotation error.}
To simulate potential errors that may arise during manual annotation, we conduct experiments with varying click annotation ranges, as detailed in Table \ref{tab:click}. These results demonstrate that our method exhibits strong stability across different annotation ranges. Notably, even as the click range expands to 0.5 meters, there is no significant drop in performance, with both mAP and mIoU remaining well within acceptable limits. This robustness indicates that our approach is resilient to variations in the click annotation radius, consistently maintaining high performance regardless of the distance from the instance center. Such stability is crucial in real-world applications, where manual annotations can introduce variability, yet reliable segmentation results are still required.
\begin{table}[t]
    \centering
    \adjustbox{max width=0.6\linewidth}{
    \begin{tabular}{ccc}
    \toprule
    Error Range (m) & mAP            & mIoU             \\ 
    \hline
    0.0         & \textbf{55.35} & 74.770           \\
    \hline
    0.1         & 54.86          & 75.688           \\
    0.2         & 54.98          & 75.591           \\
    0.3         & 54.39          & \textbf{75.936}  \\
    0.5         & 54.76          & 74.444           \\
    \bottomrule
    \end{tabular}
    }
    \caption{\textbf{Ablation Study of the Manual Annotation Error.} The first column of the table presents the click error range during manual annotation. Single click annotations are randomly selected from a circular region centered at the instance's center in the BEV plane, with the \textit{Error Range} defining the radius. A value of 0.0 indicates that the point closest to the instance center is selected.}
    \label{tab:click}
    \vspace{-3mm}
\end{table}

\noindent\textbf{Fine-tuning with the YoCo.}
Following ~\cite{jiang2024mwsis}, we fine-tune the trained network within our YoCo framework across different proportions of fully supervised data. As shown in Figure \ref{fig:finetune}, fine-tuning with a mere 0.8\% of the fully supervised data achieves performance comparable to that obtained with full supervision (59.27\% vs. 59.26\%). Moreover, when increasing the fully supervised data utilization to 5\%, our approach surpasses the fully supervised performance of the state-of-the-art PTv3.


The above experiments demonstrate that our YoCo framework achieves performance comparable to fully supervised methods using only a minimal amount of annotations, significantly reducing annotation costs.

\begin{table}[ht]
    \centering
    \adjustbox{max width=1.0\linewidth}{
    \begin{tabular}{ccccc} 
    \toprule
    Supervision                & Annotation    & Model         & YoCo         & mAP    \\ 
    \hline
    \multirow{3}{*}{Full}      & 3D Mask       & Cylinder3D~\cite{zhu2021cylindrical}   & $\mbox{-}$   & 51.40  \\
                               & 3D Mask       & SparseUnet~\cite{shi2020PartA2}   & $\mbox{-}$   & 59.26  \\
                               & 3D Mask       & PTv3~\cite{wu2024ptv3}         & $\mbox{-}$   & 60.08  \\ 
    \hline
    \multirow{6}{*}{Weak}      & Click$^\dag$  & Cylinder3D~\cite{zhu2021cylindrical}   & $\mbox{-}$   & 45.12  \\
                               & Click$^\dag$  & SparseUnet~\cite{shi2020PartA2}   & $\mbox{-}$   & 47.37  \\
                               & Click$^\dag$  & PTv3~\cite{wu2024ptv3}          & $\mbox{-}$   & 46.59  \\ 
    \cline{2-5}
                               & Click$^\dag$  & Cylinder3D~\cite{zhu2021cylindrical}   & $\checkmark$ & 49.71  \\
                               & Click$^\dag$  & SparseUnet~\cite{shi2020PartA2}   & $\checkmark$ & 55.35  \\
                               & Click$^\dag$  & PTv3~\cite{wu2024ptv3}          & $\checkmark$ & 53.83  \\ 
    \hline
    \multirow{3}{*}{Fine-Tune} & 0.8\% 3D Mask & Cylinder3D~\cite{zhu2021cylindrical}& $\checkmark$ & 53.13  \\
                               & 0.8\% 3D Mask & SparseUnet~\cite{shi2020PartA2}& $\checkmark$ & 59.27  \\
                               & 5.0\% 3D Mask   & PTv3~\cite{wu2024ptv3}        & $\checkmark$ & 60.44  \\
    \bottomrule
    \end{tabular}
    }
    \caption{\textbf{Generality Experiment of the YoCo.}}
    \label{tab:yoco4other}
    \vspace{-3mm}
\end{table}

\noindent\textbf{The generality of the YoCo.}
In Table \ref{tab:yoco4other}, we present experimental results on training various networks, including Cylinder3D, SparseUnet, and PTv3, within the YoCo framework to validate the generality of our approach. As shown, the YoCo exhibits strong performance across all three networks under weak supervision. Specifically, for the 3D instance segmentation task, our method improves mAP by 4.59\% for Cylinder3D, 7.24\% for SparseUnet, and 7.98\% for PTv3. In addition, we fine-tune different networks trained under the YoCo framework using a small portion of fully supervised data. The results indicate that with only 0.8\% of fully supervised data for fine-tuning, both Cylinder3D and SparseUnet surpass their fully supervised performance. When 5\% of labeled data is used, the PTv3 network also exceeds its fully supervised performance.

These results reveal that YoCo is not only effective for a single network architecture but also adaptable to multiple architectures, demonstrating its generality.
\section{Conclusion}
In this paper, we introduce YoCo, a novel framework for LiDAR point cloud instance segmentation using only click-level annotations. YoCo aims to minimize the performance gap between click-level supervision and full supervision. We achieve this through two key components: the VFM-PLG module, which generates high-quality pseudo labels using the zero-shot capability of the VFM model combined with geometric constraints from the point cloud, and the TSU and ILE modules, which refine labels online and offline, leveraging neural network robustness and generalization.
Our extensive experiments demonstrate that YoCo not only outperforms previous weakly supervised methods but also surpasses fully supervised methods based on the Cylinder3D, significantly reducing labeling costs while maintaining high segmentation performance. Additionally, our framework exhibits strong generality, making it applicable to various networks. These results highlight the efficiency and robustness of our approach, offering a practical solution for reducing annotation overhead in large-scale point cloud segmentation tasks.


{
    \small
    \bibliographystyle{ieeenat_fullname}
    \bibliography{main}
}
\clearpage
\setcounter{page}{1}
\maketitlesupplementary

\section{Appendix}
\label{sec:appendix}
Our supplementary material contains the following details:
    \begin{itemize}
        \item \textbf{Algorithm Implementation.} In Section \ref{sec:tsu_alg}, we provide algorithmic implementations of the TSU.
        \item \textbf{More Experiments.} In Section \ref{sec:more experiments}, we discuss the impact of YoCo on the field of autonomous driving.
        \item \textbf{Visualization.} In Section \ref{sec:visualization}, we provide some visualization results.
    \end{itemize}

\subsection{Algorithm Implementation}
\label{sec:tsu_alg}

In Algorithm \ref{alg:tsu}, we provide the detailed implementation of the TSU module, omitting the voxelization process of the point cloud.

\begin{algorithm}[ht]
    \SetAlgoLined 
	\caption{TSU}
    \label{alg:tsu}
	\KwIn{
 
    $s_v$ is voxel size. 

    $D$ is the depth threshold.

    $S$ is the voxel voting space.

    $Y$ is the current frame pseudo label. 
    
    $T_{v}$ is the threshold for the number of votes. 
    
    $T_{s}$ is the threshold for the confidence scores. 

    $(v_{x_e}, v_{y_e}, v_{z_e})$ is the ego voxelization coordinates.
 
    $V_a=\{v_a:\{(x_a, y_a, z_a, s_a)\}_i^{n}\}$ is adjacent frame point voxelization dictionary, where $v_a$ is the set 
    
    of $n$ points whose current voxelization coordinates are $(v_{x_a},$ $v_{y_a}, v_{z_a})$, and each set contains the coord- inates $(x_a, y_a, z_a)$ and confidence scores $s_a$ for the current voxel $v_a$. 
 
    $V_t=\{v_t:\{$ $(x_t, y_t, z_t, s_t)\}_i^{n'}\}$ is current frame po- int voxelization dictionary. 
    }
	\KwOut{
    
    Updated pseudo label $\hat{Y}$.}
        
	\For{$v_a$ \text{in} $V_a$}{
        \tcp{1.Update Threshold.}
		$dist = s_v\sqrt{(v_{x_a}-v_{x_e})^2+(v_{y_a}-v_{y_e})^2} $\;
        $T'_{vote} = {dist}/{D}\times T_{vote}$\;
        $T'_{score} = {dist}/{D}\times T_{score}$\;
        \tcp{2.Build Voxel Voting Space.}
        $\bar{s}=\frac{1}{n}\left(\sum{s_a}\right)$\;
		\If{$\max\left(\bar{s}\right)\ge T'_{score}$ \text{and} $n\ge T'_{vote}$,}{
			$S\left[v_{x_a}, v_{y_a}, v_{z_a}\right]=\arg\max_{cls}\left(\bar{s}\right)$;
		}
        \Else{$S\left[v_{x_a}, v_{y_a}, v_{z_a}\right]=-1$;}
	}

        \tcp{3.Update Current Pseudo Label.}
        $\hat{Y} = S\left[v_{x_t}, v_{y_t}, v_{z_t}\right]$\;
        $Mask = \hat{Y}==-1$\;
        $\hat{Y}\left[Mask\right]=Y\left[Mask\right]$.
\end{algorithm}

\subsection{More Experiments}
\label{sec:more experiments}
\begin{figure}[ht]
    \centering
    \includegraphics[width=1.0\linewidth]{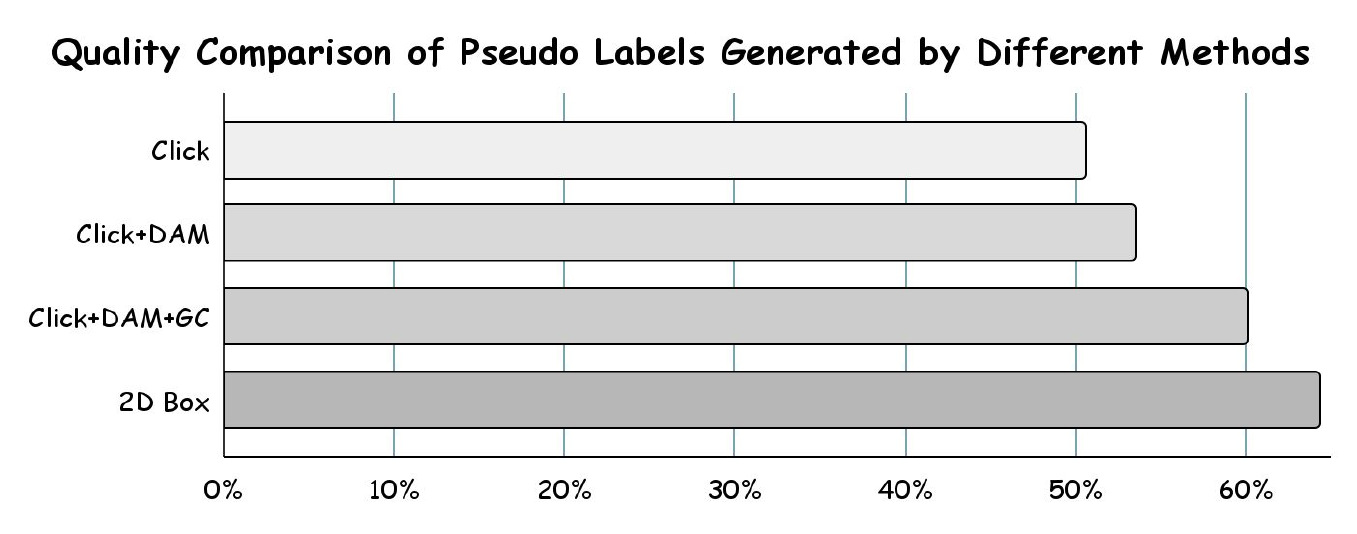}
    \caption{\textbf{Comparisons of IoU on WOD Validation Dataset.} We combine click-level and 2D bounding box-level annotations with various methods to generate pseudo labels for the validation dataset and evaluate them with ground truth.}
    \label{fig:miou}
\end{figure}

\noindent\textbf{Pseudo label quality comparison.}
Figure \ref{fig:miou} demonstrates the effectiveness of our method in generating high-quality pseudo labels, and our VFM-PLG module achieves nearly a 10\% performance improvement compared to the method that uses only clicks as prompts. 

\noindent\textbf{Results on nuScenes.} We conduct experiments on the nuScenes dataset, using only the point clouds projected onto the image. Data augmentation for single-frame data included global rotation, translation, scaling, random flipping, and shuffling. We specifically focus on the barrier, bicycle, bus, car, construction vehicle, motorcycle, pedestrian, traffic cone, trailer, and truck categories for experiments. As shown in Table \ref{tab:nuscenes}, our method YoCo achieves a 14\% improvement in mAP and a 17.868\% improvement in mIoU compared to the Click-based method, demonstrating its effectiveness.
\begin{table}[ht]
    \centering
    \adjustbox{max width=0.8\linewidth}{
    \begin{tabular}{ccccc} 
    \hline
    Sup.                  & Method               & mAP            & mIoU            & mIoU (\textit{fg})         \\ 
    \hline
    \multirow{3}{*}{Full} & SparseUnet           & \textbf{49.56} & 68.056          & 78.548           \\
                          & Cylinder3D           & 45.74          & 66.198          & 76.335           \\
                          & PTv3                 & 46.1           & \textbf{71.972} & \textbf{81.264}  \\ 
    \hline
    \multirow{2}{*}{Weak} & Click                & 19.35          & 37.686          & 58.067           \\
                          & \textbf{YoCo (Ours)} & \textbf{33.35} & \textbf{55.554} & \textbf{68.267}  \\
    \hline
    \end{tabular}
    }
    \caption{\textbf{Performance Comparisons on NuScenes Validation Dataset.} \textit{fg} denotes that only foreground points participate in the evaluation, ignoring background points.}
    \label{tab:nuscenes}
    \vspace{-2mm}
\end{table}

\noindent\textbf{Greater click range error.}
To evaluate the robustness of YoCo for larger target sizes in autonomous driving, we conduct additional experiments with error ranges of 1.0m, 1.5m, and 2.0m for vehicle classes, as shown in Table \ref{tab:veh_range}. Expanding the click range to 2.0m results in stable performance and even improvement, demonstrating the robustness of our method. These experiments suggest that the BEV center is not the optimal click, and identifying the optimal click will be explored in future work.
\begin{table}[ht]
    \centering
    \adjustbox{max width=1.0\linewidth}{
    \begin{tabular}{ccccccc} 
    \hline
    Range & 0.0    & 0.1    & 0.5    & 1.0    & 1.5    & 2.0     \\ 
    \hline
    AP   & 67.69  & 67.28  & 67.70  & 67.22  & 67.60  & \textbf{68.06}   \\
    IoU  & 81.136 & 82.288 & 82.229 & 84.350 & \textbf{85.400} & 83.885  \\
    \hline
    \end{tabular}
    }
    \caption{\textbf{Greater Click Range Error for Vehicle.}}
    \label{tab:veh_range}
    \vspace{-4mm}
\end{table}

\noindent\textbf{False positives (FP) click.}
We consider that some small objects (such as \textit{Construction Cone/Pole}) may be mislabeled as people in the annotation. To address this issue, we propose incorporating 2D semantic information to improve segmentation accuracy. Specifically, we intentionally introduce \textit{Construction Cone/Pole} objects as \textit{Pedestrian} noise in each scene to simulate real-world mislabeling scenarios. As demonstrated in Tab. \ref{tab:box}, our method achieves 53.48\% mAP and 74.936\% mIoU when using ground truth (GT) boxes, which is comparable to the performance of YoCo. Furthermore, even with imprecise predictions from YOLO, our approach maintains robust performance, validating the efficacy of leveraging 2D semantic information for verification.
\begin{table}[ht]
    \vspace{-4mm}
    \centering
    \adjustbox{max width=0.4\linewidth}{
    \begin{tabular}{cccc} 
    \hline
    Box   & GT    & YOLO  \\ 
    \hline
    mAP   & \textbf{55.12} & 53.48      \\ 
    \hline
    mIoU  & \textbf{75.120} & 74.936      \\
    \hline
    \end{tabular}
    }
    \caption{\textbf{YoCo with FP Problem.}}
    \label{tab:box}
    \vspace{-4mm}
\end{table}

\noindent\textbf{Discard click analysis.}
To verify whether all foreground objects need to be clicked, we conduct a random discard experiment. It is noteworthy that our default method inherently discards objects with insufficient point cloud representations in the Bird's Eye View (BEV) perspective, resulting in a 16.31\% object discard rate. We further evaluate the impact of additional discarding by randomly removing instances with probabilities of 10\%, 25\%, and 30\%, as detailed in Table \ref{tab:drop1} (b), (c), and (d). The experimental results indicate that as the discard rate increases, the network performance gradually declines.  However, even with a discard proportion of 37.19\%, our method still achieves 53.96\% mAP and 73.248\% mIoU, demonstrating the robustness of YoCo.

\begin{table}[ht]
    \centering
    \adjustbox{max width=0.8\linewidth}{
    \begin{tabular}{ccccc} 
    \hline
        & Method                                                                   & Proportion     & mAP            & mIoU             \\ 
    \hline
    (a) & YoCo                                                                     & 16.31\%        & \textbf{55.35} & \textbf{74.770}  \\ 
    \hline
    (b) & \multirow{3}{*}{\begin{tabular}[c]{@{}c@{}}Discard\\Object\end{tabular}} & 24.65\% (10\%) & 54.29          & 73.742           \\
    (c) &                                                                          & 37.19\% (25\%) & 53.96          & 73.248           \\
    (d) &                                                                          & 58.10\% (50\%) & 46.68          & 70.008           \\
    \hline
    \end{tabular}
    }
    \caption{\textbf{Ablation Study of Drop Click.}}
    \label{tab:drop1}
    \vspace{-2mm}
\end{table}

\noindent\textbf{Scalability across sequences.}
In Table \ref{tab:drop2}, we assess the efficacy of YoCo under limited annotation scenes by subsampling point cloud sequences at intervals. Specifically, in Table \ref{tab:drop2} (b) and (c), we report performance metrics using 1/3 and 1/2 of the annotated sequences, respectively. Notably, even with only 1/3 of the annotation data, our method achieves 53.13\% mAP and 72.304\% mIoU, highlighting its scalability and robustness in resource-constrained settings.
\begin{table}[ht]
    \centering
    \adjustbox{max width=0.55\linewidth}{
    \begin{tabular}{cccc} 
    \hline
        & Interval & mAP            & mIoU             \\ 
    \hline
    (a) & 1        & \textbf{55.35} & \textbf{74.770}  \\ 
    \hline
    (b) & 2        & 53.90          & 72.999           \\
    (c) & 3        & 53.13          & 72.304           \\
    \hline
    \end{tabular}
    }
    \caption{\textbf{Ablation Study of Across Sequences.}}
    \label{tab:drop2}
    \vspace{-2mm}
\end{table}

\noindent\textbf{Unsupervised application of YoCo.}
To extend the applicability of our approach, we investigate the use of the 2D detection model YOLO for unsupervised point cloud segmentation tasks. Specifically, we align YOLO's detection categories with the annotation categories of the Waymo dataset. Since YOLO does not directly support the \textit{Cyclist} category, we propose a heuristic strategy: the \textit{Cyclist} category is inferred by combining detections from the intersection of the \textit{Person} and \textit{Bicycle} categories. As shown in Table \ref{tab:uns}, our unsupervised approach achieves significant performance improvements, with a 5.59\% increase in mAP and a 4.672\% increase in mIoU compared to the click-based method. These results validate the effectiveness of leveraging 2D detection models for unsupervised point cloud segmentation tasks.
\begin{table}[ht]
    \centering
    \adjustbox{max width=0.95\linewidth}{
    \begin{tabular}{ccccc} 
    \hline
    Supervision           & Annotation & Model      & mAP   & mIoU    \\ 
    \hline
    Full                  & 3D Mask    & SparseUnet & 59.26 & 79.505  \\ 
    \hline
    \multirow{2}{*}{Weak} & Click$^*$  & SparseUnet & 40.19 & 67.510  \\
                          & Click$^\dag$ & YoCo       & 55.35 & 74.770  \\ 
    \hline
    Unsupervised          & YOLO       & YoCo       & 45.78 & 72.182  \\
    \hline
    \end{tabular}
        }
     \caption{\textbf{Performance Comparison of Supervision Strategies on the Waymo Validation Dataset.} YOLO refers to pseudo labels derived from YOLO prediction results.}
    \label{tab:uns}
\end{table}

\subsection{Visualization}
\label{sec:visualization}
\begin{figure*}[t]
\centering
   \includegraphics[width=0.95\linewidth]{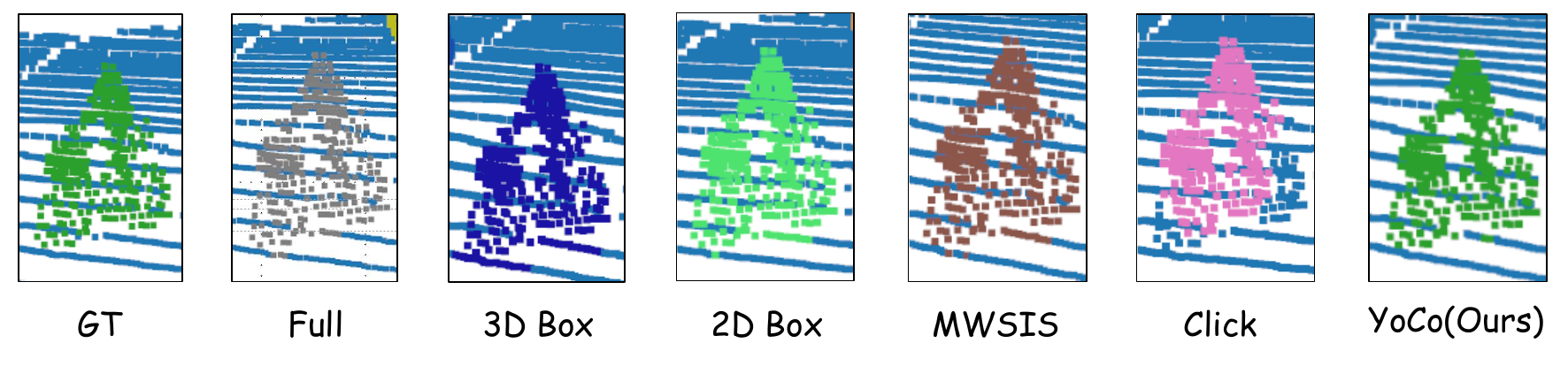}
   \caption{\textbf{Visiualization for Cyclist.}}

\label{fig:vis_cyc}
\end{figure*}

In Figure \ref{fig:vis_cyc}, we evaluate the performance of SparseUnet trained with various annotation strategies. Our proposed method exhibits superior visual quality, significantly outperforming click-based approaches by generating more accurate predictions. This improvement is largely driven by the integration of deep prior information into the pseudo label generation process, which enhances the quality of the generated labels.

\begin{figure*}[t]
\centering
   \includegraphics[width=1.0\linewidth]{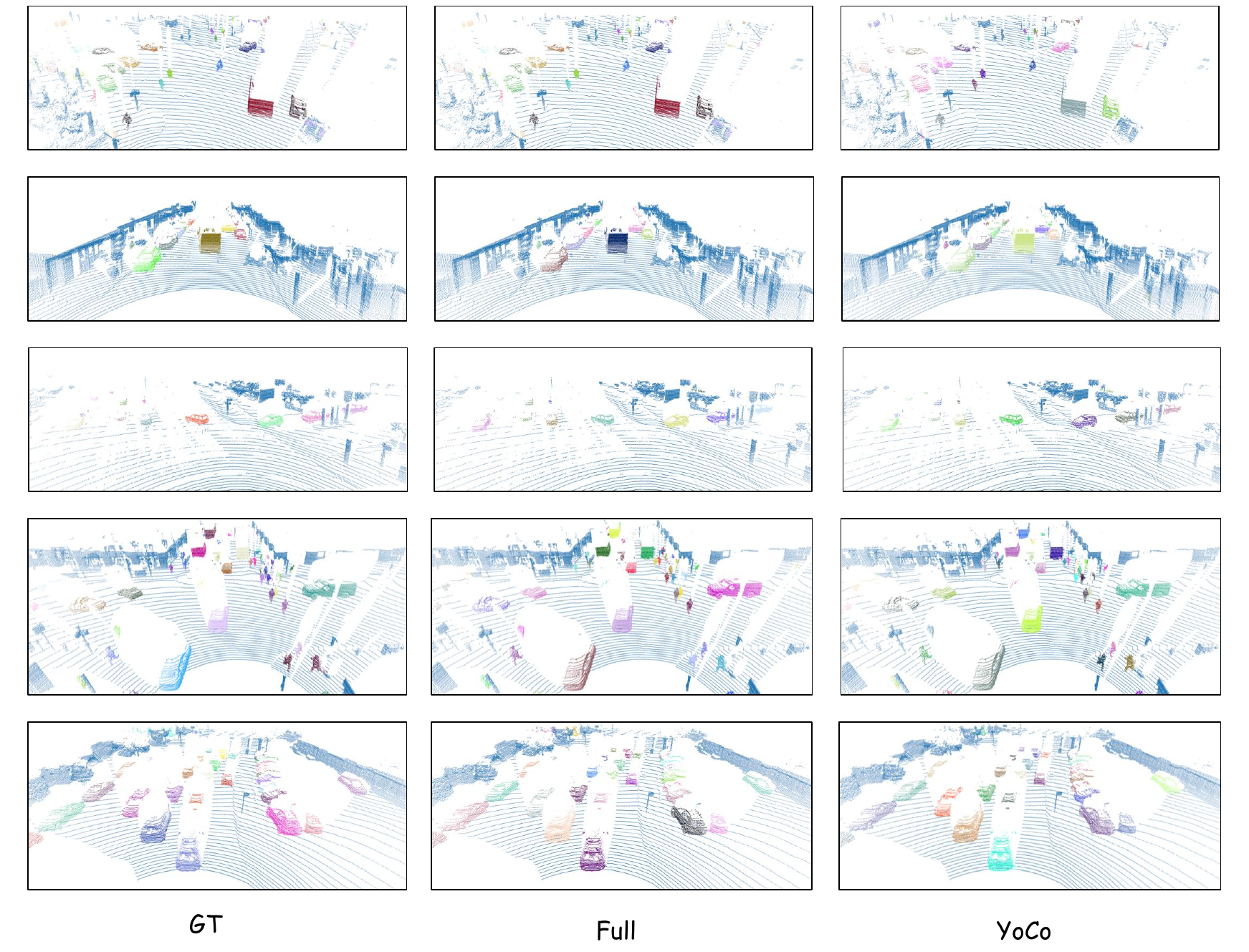}
   \caption{\textbf{Visiualization for Scenes.}}

\label{fig:vis_scene}
\end{figure*}

In Figure \ref{fig:vis_scene}, we provide visualization segmentation results for some scenes.

\end{document}